# DESIGN AND INTEGRATION OF END-EFFECTOR FOR 3D PRINTING OF NOVEL UV-CURABLE SHAPE MEMORY POLYMERS WITH A COLLABORATIVE ROBOTIC SYSTEM

Luis Velazquez, Genevieve Palardy, Corina Barbalata

Department of Mechanical & Industrial Engineering, Louisiana State University

3261 Patrick F. Taylor Hall

Baton Rouge, LA 70803

## ABSTRACT

This paper presents the initial development of a robotic additive manufacturing technology based on ultraviolet (UV)-curable thermoset polymers. This is designed to allow free-standing printing through partial UV curing and fiber reinforcement for structural applications. The proposed system integrates a collaborative robotic manipulator with a custom-built extruder end-effector designed specifically for printing with UV-curable polymers. The system was tested using a variety of resin compositions, some reinforced with milled glass fiber (GF) or fumed silica (FS) and small-scale, 2D and 3D specimens were printed. Dimensional stability was analyzed for all formulations, showing that resin containing up to 50 wt% GF or at least 2.8 wt% FS displayed the most accurate dimensions.

## 1. INTRODUCTION

Additive Manufacturing (AM), the process of creating 3D objects through deposition of material in a layer-by-layer fashion, has become commonplace in the aerospace and automobile industries, as well as in various research areas. It allows to manufacture complex shape objects relatively quickly through the integration of automated machines and Computer Aided Design (CAD) software [5]. Furthermore, there are different AM methods that can achieve good dimensional accuracy in the finished product using multiple types of materials, such as thermoplastics, metal powder, UV-curable resins, wood, adhesives, and others. While this manufacturing method has not replaced traditional ones like injection molding or machining, it has slowly been incorporated in different industries to create prototypes or investigate the dimensional accuracy of printed parts [6]. More recently, there has been an increase in research regarding the mechanical properties of 3D printed samples in academic settings. For example, the potential use of the technology in the medical field to produce bone or tissue scaffolds is being explored [5]. In addition, the last decade has seen increased popularity of Fused Deposition Modeling (FDM) 3D printers among non-academic circles as a segment of the population adopt AM as a hobby [6]. FDM 3D printers are easy to use from home and they typically consist of a compact machine with a heated extrusion system. The extruder is fed a polymer filament, melts it, and deposits it on a platform to create 3D objects.

Related research in this area has also combined AM technology to print with photopolymers. For example, Direct Ink Writing (DIW) can produce 3D printed samples using UV-curable resins with specially designed printing heads [10, 11]. In these approaches, electromechanical controlled extrusion of UV-curable composite filament is combined with UV light exposure that follows the extrusion point. This design demonstrated new prospects for the manufacturing of complex 3D shapes using composites. In other cases, traditional FDM machines have been modified to include extrusion systems capable of 3D printing self-supported structures using photocurable resins [16-17].

The benefits of using robots in the manufacturing industry are well recognized. They facilitate consistent production rates, increase capacity, improve product quality, and reduce waste. Specifically, robotic additive manufacturing has several advantages over traditional, in-plane layer-by-layer 3D printing approaches such as large-scale printing, extensive mobility, cost effectiveness, and increased part complexity. Furthermore, leveraging the high dimensionality of articulated industrial manipulators, complex parts can be printed by following convoluted paths along the direction of curved surfaces. In the past years, companies such as Continuous Composites [13] (ID, USA) and Moi Composites (Italy) [14] started looking into large-scale, extrusion AM robotic systems for reinforced continuous fiber polymers to enable free-standing printing and design of complex, integrated geometries, while eliminating the need for tooling and autoclaves for composites manufacturing. Other developments in the field have been achieved by Massivit (Isreal) [18] and Mighty Buildings (CA, USA) [19]. These companies are capable of printing fully cured structures for a variety of applications in the construction, entertainment, aerospace, and education industries, while relying on commercial large scale traditional 3D printing machines.

This paper presents initial work to demonstrate the feasibility of collaborative robotic manipulators to achieve 3D printing using UV-curable resins. Equipped with custom-made end-effectors, collaborative manipulators can enable autonomous 3D printing of complex shapes in close proximity to humans, ensuring safe human-robot cohabitation. The proposed system will allow free-standing printing of specimens as the end-effector is specifically designed to print with UV-curable resins. Furthermore, the long-term goal of this technology is to automate the 3D printing process with novel TSMPs as part of the larger Louisiana Materials Design Alliance (NSF RII Track-1: LAMDA) project.

The paper will focus on the integration of the UR5e collaborative arm with a custom-built end-effector extrusion system and assessment of its potential for additive manufacturing of UV-curable thermosets. The proposed design is centered around commercially available UV-curable polymers to evaluate some of the most common problems in material deposition related to viscosity or curing behavior. UV light sources were attached to the extruder to cure the resin as it was deposited, and resins with different filler formulations were explored to 3D print small-scale specimens. Figure 1(a) shows the custom-designed end-effector attached to the robotic manipulator.

The rest of the paper is structured as follows: in Section 2, the materials used for this project and the robotic extruder head development and integration are described; in Section 3, preliminary results for 3D printed specimens and a discussion regarding their dimensional accuracy are presented, followed by main conclusions in Section 4.

## 2. METHODOLOGY

### 2.1 Extruder design

The extruder designed for this research must have the following capabilities: 1) controllable material deposition, 2) easy integration with robotic manipulators, 3) adaptable to different UV-curable TSMP materials, and 4) affordable.

The designed extruder, seen in Figure 1(a), consists of a syringe barrel containing up to 200 mL of uncured resin fitted with an extrusion nozzle that has a defined diameter of 1.5 mm. The material is extruded through the nozzle by applying a predefined pressure and the deposited material is exposed to an UV light (10 W, 365 nm wavelength spotlight). The extrusion rate of the deposited material stayed constant at approximately 5.3 $mm^3$/s. To ensure control of the material deposition and apply pressure, a stepper motor was integrated with the system. The NEMA 23 was chosen because of its higher torque of 1.9 Nm, which is needed to drive the plunger against internal friction and high viscosity resins. Using an Arduino UNO and a DM556T stepper driver, the speed and revolutions of the motor shaft were controlled to obtain the desired flow rates to print adequately. Furthermore, this stepper motor can be easily integrated with the command unit of the robotic manipulator, allowing for coordinated control of both systems.

Various parts of the extruder, such as the attachment to the robotic arm, the enclosure of the extruder, the press-fit ring of the UV light source, and the plunger, were 3D printed using FDM printers and PLA filament. Although PLA has lower strength than metal, having the main parts 3D printed allows for easy modifications as needed throughout the design process of the prototype. Figure 2 shows the SolidWorks models of the main components of the extruder and an assembled model.

The prototype was assembled as shown in Figure 1(a). The syringe body of the extruder was adapted to accept different nozzles. For the experiments presented in this paper we used a 1.5 mm nozzle. The UV spotlight was attached to the syringe using a press-fit at the tip of the syringe barrel, which directs the light at an angle of 24 degrees approximately as shown in Figure 1(b). A light blocking wall had to be introduced between the nozzle and the UV spotlight to prevent curing inside of the nozzle.

The stepper motor was coupled with a lead screw to drive a plunger up or down when refilling the syringe with resin or depositing it, respectively. Closed-loop regulators are designed to control the extrusion rate of the material. This end-effector features an external resin feeding line, eliminating the need for syringe refilling and reducing time during printing. The extruder is mounted through the adaptor seen in Figure 2(a) to the UR5e collaborative manipulator.

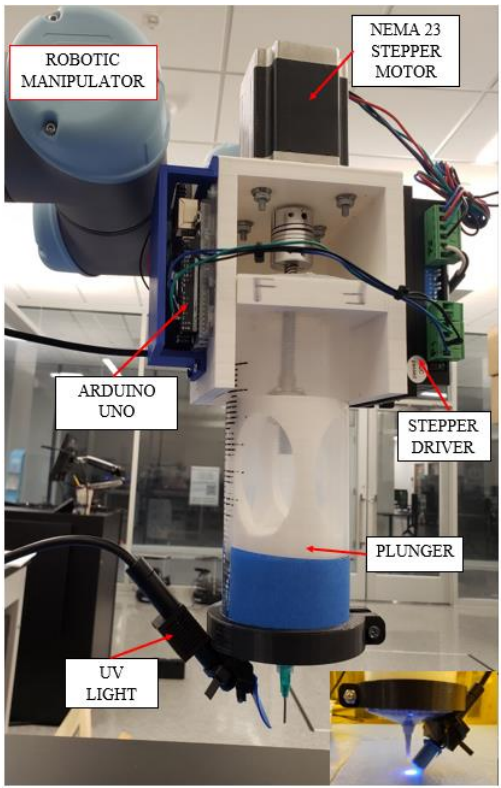
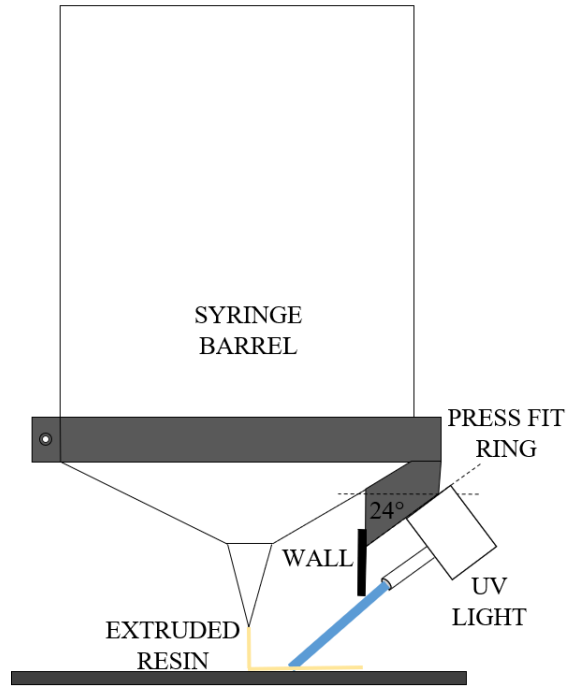

(a) End- effector attached to a UR5e manipulator.

(b) Schematic of the UV light and syringe barrel parts of the extrusion system.

Figure 1. Robotic UV-curable resin extrusion system.

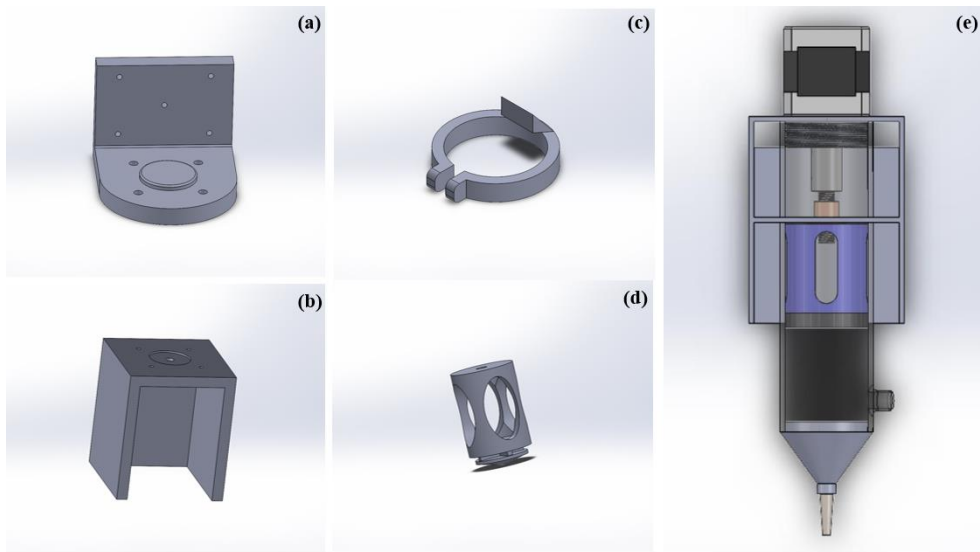

Figure 2. SolidWorks models of the main components of the extruder, a) Attachment to the robotic arm, b) Enclosure of the extruder, c) Press-fit ring of the UV light source, d) Plunger, and e) Assembled extruder prototype.

## 2.2 Robotic arm control

UR5e robotic arm is a collaborative manipulator that adapts its behavior to a dynamic environment. Specifically, the robotic arm stops when a collision is detected, ensuring that humans are not harmed or objects are not damaged. The robotic arm has six degrees-of-freedom, allowing full control of its end-effector.

A motion planning algorithm based on the CAD model of the printed part was developed for the robotic arm. This controls the velocities of the manipulator and ensures coordination with the material extrusion rate, providing accurate deposition of the material.

The flow chart in Figure 3 summarizes the steps taken to control the robotic arm and print various samples. The process includes the design of the CAD representation of the printed sample, the generation of the path the extruder must take to print the sample, the transformation of the path considering the environment and the robot configuration, the printing of the sample, and the qualitative and quantitative evaluation of the print.

A CAD representation of the printed specimen was designed using SolidWorks. Next, to achieve 3D printed specimens, the desired path of the end effector was sketched in SolidWorks. From here, the Universal Robots (UR) add-on was used to create the g-code needed for the extrusion path. Figure 4(a) shows an example of sketch made in Solidworks to print a rectangle (90 mm x 60 mm). Due to the configuration chosen for the UV light with respect to the extruder nozzle, it was required to add 25 mm-long sections to the original dimensions of the rectangle to account for the distance between the UV wall and the nozzle tip. This ensured that all sections, including the corners of the rectangle, were cured evenly.

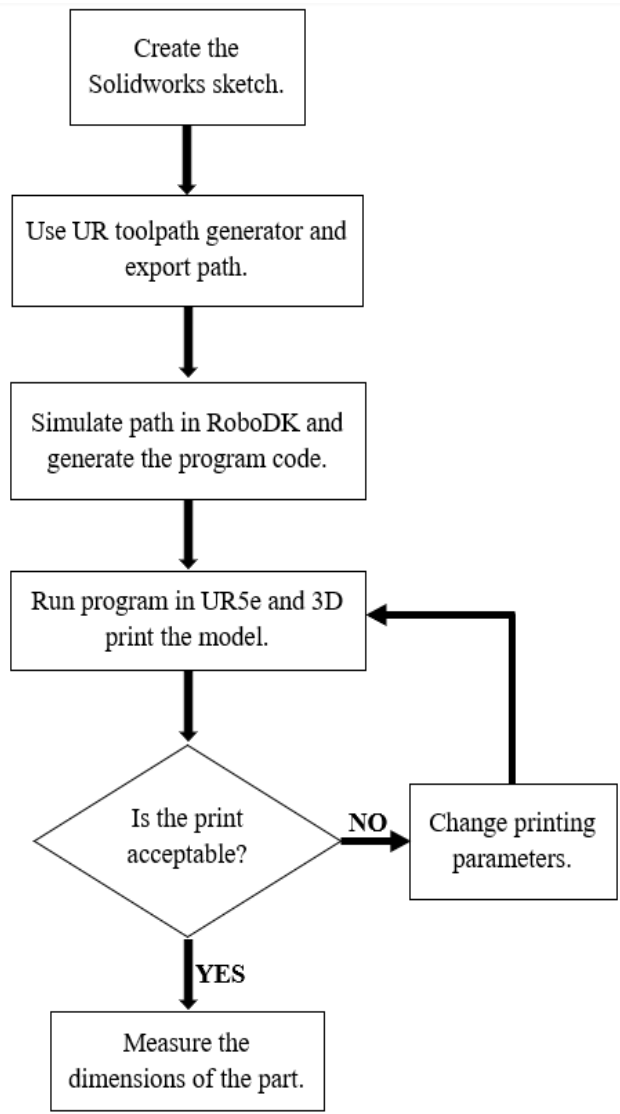

Figure 3. 3D printing process flowchart using a URe5 robotic manipulator.

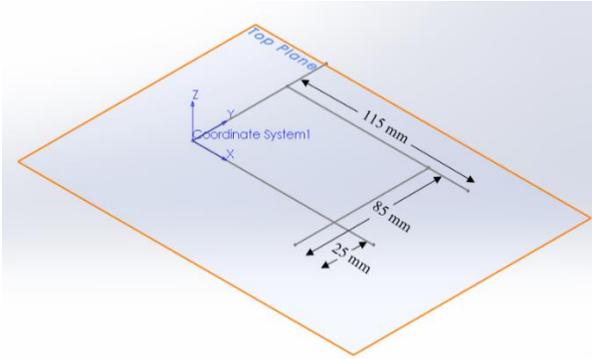
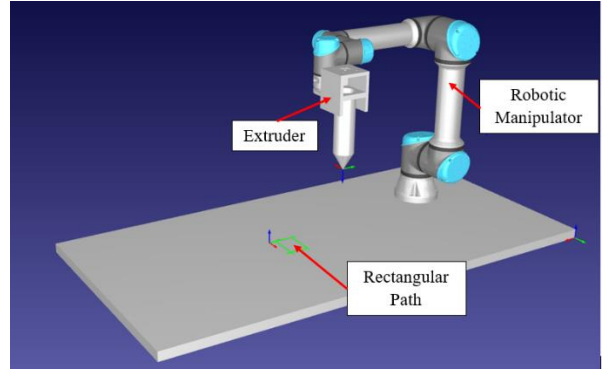

(a) SolidWorks sketch of the 3D printed specimen (90 mm x 60 mm rectangle).

(b) RoboDK simulation environment.

Figure 4. Robotic arm path generation and simulation.

Up to now, the extruder's path was created without any considerations regarding the environment where it operates or the configuration of the robot. To address these limitations, we used the RoboDK software [15], to generate the robot motion ensuring accurate and safe prints. Furthermore, RoboDK was used to simulate the 3D printing before the program is run in the robotic manipulator. This makes sure no errors or collisions happen between the real robot and the working environment. First, the CAD model of the newly designed end-effector is integrated with the UR5e robot simulation. RoboDK allows to change the orientation of the end-effector for each printing path to guarantee there are no collisions with the table or other structures.

Then, to create the code that simulates printing of a 3D object, the model's g-code was imported into RoboDK. By doing this, the path the extruder head must follow is displayed in the working environment. From here, the software modifies different process parameters as desired, such as end-effector orientation, position, and speed to ensure no collision with the environment. Because the UV light is fixed with respect to the nozzle, special care was given to the orientation of the end-effector to ensure curing of the resin. Figure 4(b) shows the RoboDK simulation environment to 3D print the rectangular sample shown in Figure 4(a). Once the motion of the robot is simulated in RoboDK and it is ensured that there is no collision between the robotic arm and the working environment, the code is uploaded on the UR5e manipulator, and the printing process begins. The current implementation integrates the extruder and UV lights with the controller box of the robot, offering the convenience and higher precision from automating the process. Program codes created in RoboDK by the user turn the extrusion process on and off through a series of digital signals to the Arduino and other components of the system.

## 2.3 Materials

Two commercially available UV-curable resins were studied in this research. The first one was an acrylate-based resin (urethane acrylate and 2-hydroxyethyl methacrylate), typically used for casting and molding. Its chemical formulation caused partial curing in less than one second, which created brittle specimens. The second one was an UV curing DLP resin (Anycubic), sensitive to UV wavelengths between 355 and 405 nm. It generated samples with lower degree of cure compared to the acrylic ones under the same conditions. In addition, 0.8 mm milled glass fiber (GF, Fiberglass Supply Depot Inc.) and fumed silica (FS, VWR International) were mixed into the DLP resin with different weight fractions (0, 9, 35 and 50 wt%) to study how it influenced the printing process and dimensional stability of the printed parts.

**2.4 Specimens printing**

To evaluate the performance of the proposed system, a set of 36 experiments were performed, where the robotic arm printed 2D rectangular samples (90 mm x 60 mm), shown in Figure 4(a), 3D walls (50 mm x 10 mm) and 3D squares (30 mm x 30 mm x 8.5 mm), shown in Figure 7. Among these, 9 were printed using the DLP resin with various milled GF weight fractions, and 16 were printed with 9 wt% FS for increased viscosity. The other 11 experiments used the acrylic-based resin without any additives.

After printing, the actual dimensions of all specimens were measured with a ruler accurate to 0.5 mm or digital caliper accurate to 0.01 mm. Pictures of the samples were also used to confirm dimensions with a scale bar. Measurements of the 2D rectangle length, width, and line width were taken to compare with the SolidWorks sketch dimensions as well as the height in the case of the 3D samples. Since there was some variability in the line width of a single sample, the measurements were taken at the same location for all specimens. For consistency, all 2D samples were printed at a constant robotic arm speed of 3 mm/s and all 3D ones used a speed of 4 mm/s.

# 3. RESULTS AND DISCUSSION

**3.1 2D specimens**

**3.1.1 Specimens using DLP resin**

Milled GF was added at different weight fractions to study its effect on dimensional accuracy and curing behavior. Figure 5(a) to (c) show examples of the printed specimens with the collaborative robotic manipulator. It is observed that higher weight fractions of milled GF improved dimensional stability. Higher viscosity resin formulations achieved superior dimensional stability, since the shape of the extruded line was maintained until curing was triggered by the spotlight. Figure 6 summarizes the measured dimensions of the 2D printed samples for all milled GF and FS contents. Expected values for length and width of the rectangles were 90 mm and 60 mm, respectively. Filler content affected the dimensional accuracy of the specimens. The closest results to the original dimensions of the rectangle (90 mm x 60 mm), as seen in Figure 5(c) and Figure 6, were the samples printed with 50 wt% GF or 2.8 wt% FS. These formulations had the highest viscosity and retained the extruded filament shape before the UV light started the curing process. The resin with 35 wt% GF showed limited improvement over 0 wt% GF since its viscosity did not seem to be affected significantly.

It was visually observed that at higher percentages of milled glass fiber, the outer layer of cured resin was thicker compared to the resin without additives. A possible reason for this is that the UV light dispersed in all directions because of the glass fibers present, causing more localized curing. As reported in Figure 6, the specimens printed with the highest percentage of milled glass fiber achieved the highest dimensional accuracy. In these cases, deformation caused by cure gradients was not as noticeable as in the samples with lower GF weight fractions.

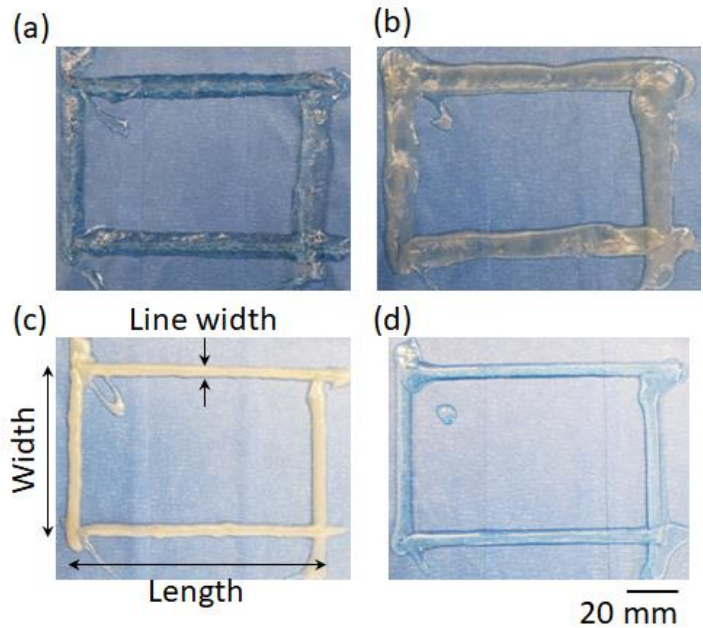

Figure 5. 2D printed specimens using UV-curable resin and a robotic manipulator. a) DLP resin with 0 wt% milled GF, b) DLP resin with 35 wt% milled GF, c) DLP resin with 50 wt% milled GF, and d) Acrylic-based resin.

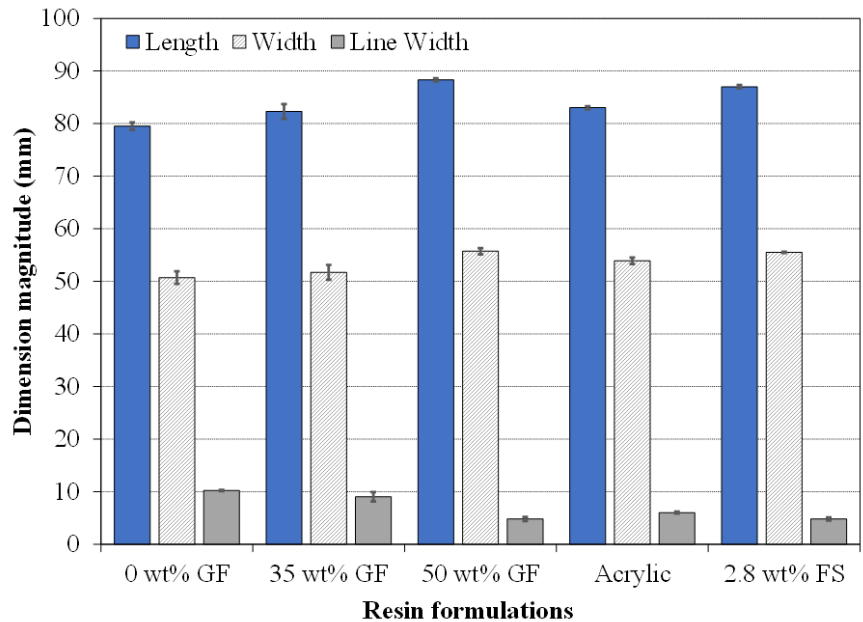

Figure 6. Dimensional analysis of 2D printed samples using UV-curable resins and a collaborative robotic manipulator.

### 3.1.2 Specimens using acrylic-based resin

Figure 6 shows the results obtained for the set of experiments using the acrylic-based resin. On average, this resin showed improved dimensional accuracy when compared to the DLP resin (with 0 wt% and 35 wt% GF).

All the 3D printed parts, as shown in Figure 5(d), were partially cured, forming a thin solid layer on the outer surface of the filaments. Specimens were brittle and exhibited deformation as the resin cured. None of the specimens fully cured through-the-thickness and liquid uncured resin remained at the bottom of the filaments, which suggests that post-curing would be required in a UV chamber. However, all samples had nearly identical shapes, indicating precision in the motion of the manipulator.

### 3.2 3D specimens with DLP resin

From the previous experiments, it was observed that curing behavior can be affected by singularities caused by the configuration of the collaborative robotic arm. For complex shapes, this would hinder the exposure to UV light for some areas on the specimens and would require manual intervention. To address this limitation, for these experiments, an additional UV light was used and mounted on the end-effector. DLP resin with 9 wt % FS was selected. Figure 7 shows a subset of the 3D walls (Figure 7(a) and (b)) and squares (Figure 7(c) and (d)) that were printed. Measured dimensions are indicated for each type of specimen. Line width was measured as shown in Figure 5(c).

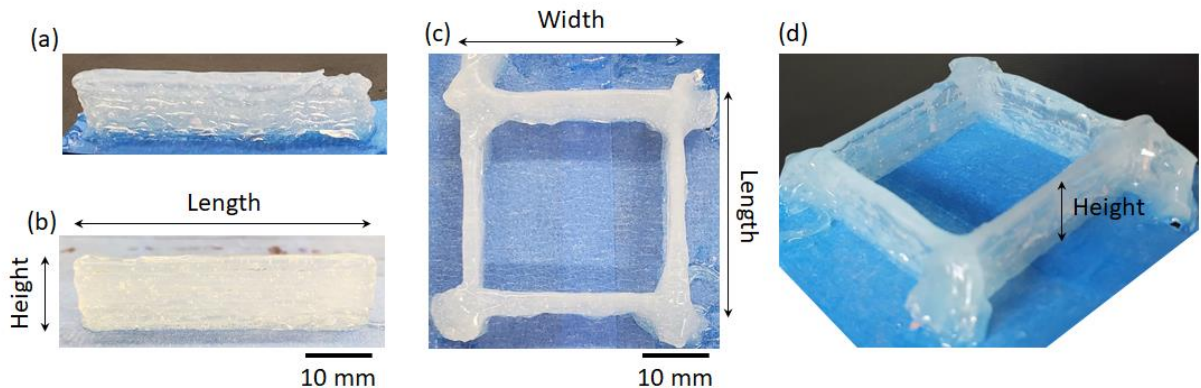

Figure 7. 3D printed walls (a and b) and squares (c and d) using UV-curable resin and the collaborative robotic manipulator.

Dimensional stability data for all 3D printed specimens is shown in Figure 8(a) for walls and Figure 8(b) for squares. On average, the walls had a width of 49.76 ± 1.27 mm (compared to 50 mm for the CAD model) and a height of 11.07 ± 0.86 mm (compared to 10 mm). For the square specimens, the average width, length and height were equal to 32.09 ± 0.11 mm, 32.01 ± 0.30 mm and 8.62 ± 0.19 mm, respectively (compared to 30 mm x 30 mm x 8.5 mm). While some values were slightly higher than those of the CAD model, they indicate the feasibility of this collaborative manipulator with custom end-effector for 3D printing.

Qualitatively, these samples exhibited higher solidification than the 2D specimens. Each layer benefited from repeated exposure to the UV light as the structure was built upwards, in addition to the second UV spotlight. This resulted in specimens that were more rigid when separated from the printing area. To fully cure the specimens, a UV chamber can be used as a post-curing method.

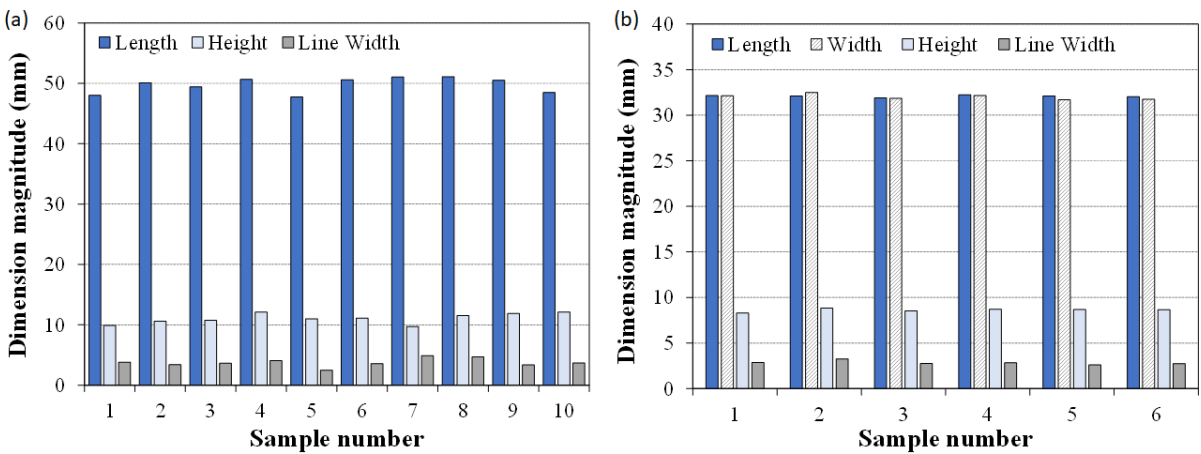

Figure 8. Measured dimensions for all 3D printed (a) wall samples (original length x height, 50 mm x 10 mm), and (b) squares (original length x width x height, 30 mm x 30 mm x 8.5 mm) using UV-curable resin with 9 wt% FS and a collaborative robotic manipulator. All measured dimensions are accurate to 0.01 mm.

## 4. CONCLUSIONS

In this work, we proposed a custom extruder design integrated with a collaborative robotic manipulator to automate the 3D printing process with UV-curable resins. We used this system to 3D print several samples and study the dimensional accuracy and curing behavior of three different resin formulations with milled GF and FS as additives to increase their viscosity.

3D printing with UV-curable resins is a significant challenge in several ways. Each resin formulation requires different printing parameter values to achieve successful results, but deformation is still highly likely to occur. Additives like milled glass fiber and fumed silica can help increase the dimensional accuracy of the specimens because it reduces curing deformation and increases viscosity. More viscous formulations (50 wt% GF and ≥ 2.8 wt% FS) performed better during the extrusion process since the extruded filament maintained its shape for longer before the UV spotlight started the curing process. In addition, the samples printed with milled glass fiber presented higher solidification when compared to other formulations.

Integrating the extrusion system with a robotic manipulator offers the possibility of automating the 3D printing process and producing highly precise prints. Moreover, it presents the starting point for more complex 3D prints, such as grid-like patterns and multi-layered specimens with intricate geometry. Future work includes viscosity and curing behavior characterization and modeling to study resin formulations suitable for free-standing 3D printing. Furthermore, other combinations of UV light intensities and robotic motions will be tested, based on dynamic cure kinetics modeling, to find the best printing parameters for different shapes and resin formulations.

## 4. ACKNOWLEDGEMENTS

This work was supported by the US National Science Foundation under grant number OIA-1946231 and the Louisiana Board of Regents for the Louisiana Materials Design Alliance (LAMDA).

## 5. REFERENCES

[1] Melly, S. K. et al. (2020) 'Active composites based on shape memory polymers: overview, fabrication methods, applications, and future prospects', Journal of Materials Science, 55(25), p. 10975. doi: 10.1007/s10853-020-04761-w.

[2] Zare, M. et al. (2019) 'Thermally-induced two-way shape memory polymers: Mechanisms, structures, and applications', Chemical Engineering Journal, 374, pp. 706–720. doi: 10.1016/j.cej.2019.05.167.

[3] "Introduction and Basic Principles." *Additive Manufacturing Technologies: Rapid Prototyping to Direct Digital Manufacturing*, by Ian Gibson et al., Springer, 2010, pp. 1–15.

[4] M. D. Vasilescu, Scientific Bulletin of Naval Academy, Vol. XXII 2019, pg. 289 - 296.

[5] Mott, Eric J, et al. "Digital Micromirror Device (DMD)-Based 3D Printing of Poly(Propylene Fumarate) Scaffolds." *Materials Science and Engineering: C*, no. 61, 2016, pp. 301–311.

[6] "CHGT: 3D Printing for the Hobby and Model Enthusiasts Market Gathering Steam." *TheNewswire.ca*, 2016. *Access World News – Historical and Current*, infoweb-newsbank-com.libezp.lib.lsu.edu/apps/news/document-view?p=WORLDNEWS&docref=news/15D1A17C0E4C9D30.

[7] Metral, B. et al. (2019) 'Photochemical Study of a Three-Component Photocyclic Initiating System for Free Radical Photopolymerization: Implementing a Model for Digital Light Processing 3D Printing', ChemPhotoChem, 3(11), pp. 1109–1118. doi: 10.1002/cptc.201900167.

[8] 'Aminothiazonaphthalic anhydride derivatives as photoinitiators for violet/blue LED-Induced cationic and radical photopolymerizations and 3D-Printing resins' (2016) Journal of polymer science, 54(9), pp. 1189–1196. doi: 10.1002/pola.27958.

[9] Hong, S. Y. et al. (2018) 'Experimental investigation of mechanical properties of UV-Curable 3D printing materials', Polymer, 145, pp. 88–94. doi: 10.1016/j.polymer.2018.04.067.

[10] Chen, K. *et al.* (2018) 'Fabrication of tough epoxy with shape memory effects by UV-assisted direct-ink write printing', *Soft matter*, 14(10), pp. 1879–1886. doi: 10.1039/c7sm02362f.

[11] Wu, T. *et al.* (2019) 'Additively manufacturing high-performance bismaleimide architectures with ultraviolet-assisted direct ink writing', *Materials & Design*, 180. doi: 10.1016/j.matdes.2019.107947.

[12] Jizhou Fan and Guoqiang Li (2018) 'High enthalpy storage thermoset network with giant stress and energy output in rubbery state', *Nature Communications*, 9(1), pp. 1–8. doi: 10.1038/s41467-018-03094-2.

[13] "Continuous Composites: Continuous Fiber 3D Printing." *Continuous Composites Continuous Fiber 3D Printing*, www.continuouscomposites.com/.

[14] Moi *Composites*, www.moi.am/.

[15] "Simulate and Program Your Robot in 5 Easy Steps." *RoboDK*, robodk.com/simulation.

[16] Mizuno, Y., Pardivala, N. and Tai, B. L. (2018) 'Projected UV-resin curing for self-supported 3D printing', *Manufacturing Letters*, 18, pp. 24–26. doi: 10.1016/j.mfglet.2018.09.005.

[17] Asif, M. *et al.* (2018) 'A new photopolymer extrusion 5-axis 3D printer', *Additive Manufacturing*, 23, pp. 355–361. Doi: 10.1016/j.addma.2018.08.026.

[18] Massivit, massivit3d.com/

[19] Mighty Buildings, mightybuildings.com/about-us